%% file: main.tex
\documentclass[9pt,twocolumn,twoside]{pnas-new}

\setboolean{displaywatermark}{false}

\input{packages.tex}

\input{macros.tex}

\graphicspath{{./figs/}}

\newif\ifarxiv
\arxivtrue %

\ifarxiv
\templatetype{preprint}
\else
\templatetype{pnasinvited}
\fi

\title{Understanding the Role of Individual Units \\ in a Deep Neural Network}

\input{text/authors.tex}


\input{text/abstract.tex}

\doi{\url{www.pnas.org/cgi/doi/10.1073/pnas.1907375117}}
\ifarxiv
\dates{This preprint was compiled on \today.  The peer-reviewed version can be found at \url{www.pnas.org/cgi/doi/10.1073/pnas.1907375117}.}
\else
\dates{This manuscript was compiled on \today.}
\fi

\begin{document}

\maketitle
\ifthenelse{\boolean{shortarticle}}{\ifthenelse{\boolean{singlecolumn}}{\abscontentformatted}{\abscontent}}{}

\input{text/introduction.tex}

\input{text/results.tex}

\input{text/applications.tex}

\input{text/discussion.tex}

\input{text/methods.tex}

\input{text/acks.tex}

\bibliography{main}

\end{document}

%% file: packages.tex
\usepackage{color,xcolor}
\usepackage{epsfig}
\usepackage{graphicx}

\usepackage{tikz}
\usetikzlibrary{arrows}

\usepackage{chngcntr}

\usepackage{tabularx}
\usepackage{adjustbox}
\usepackage{array}
\usepackage{booktabs}
\usepackage{colortbl}
\usepackage{float,wrapfig}
\usepackage{hhline}
\usepackage{multirow}

\usepackage{amsmath,amsfonts,amsthm,amssymb}
\usepackage{bm}
\usepackage{nicefrac}
\usepackage{microtype}
\usepackage{dsfont}
\usepackage{changepage}
\usepackage{extramarks}
\usepackage{fancyhdr}
\usepackage{lastpage}
\usepackage{setspace}
\usepackage{soul}
\usepackage{xspace}

\usepackage{paralist}
\usepackage{enumitem}

%% file: macros.tex
\usepackage{amsmath,amsfonts,bm}

\newcommand{\Prob}{\mathbb{P}}
\newcommand{\IoU}{\text{IoU}}

\newcommand{\reffig}[1]{Figure~\ref{fig:#1}}

\newcommand{\lblfig}[1]{\label{fig:#1}}

\newcommand{\ignorethis}[1]{}

\def\eqref#1{equation~\ref{#1}}

\def\1{\bm{1}}

\DeclareMathAlphabet{\mathsfit}{\encodingdefault}{\sfdefault}{m}{sl}
\SetMathAlphabet{\mathsfit}{bold}{\encodingdefault}{\sfdefault}{bx}{n}

\newcolumntype{L}[1]{>{\raggedright\let\newline\\\arraybackslash\hspace{0pt}}m{#1}}
\newcolumntype{C}[1]{>{\centering\let\newline\\\arraybackslash\hspace{0pt}}m{#1}}
\newcolumntype{R}[1]{>{\raggedleft\let\newline\\\arraybackslash\hspace{0pt}}m{#1}}

\newcommand{\ignore}[1]{}

\makeatletter
\DeclareRobustCommand\onedot{\futurelet\@let@token\@onedot}
\def\@onedot{\ifx\@let@token.\else.\null\fi\xspace}

\makeatother

\definecolor{MyDarkBlue}{rgb}{0,0.08,1}
\definecolor{MyDarkGreen}{rgb}{0.02,0.6,0.02}
\definecolor{MyDarkRed}{rgb}{0.8,0.02,0.02}
\definecolor{MyDarkOrange}{rgb}{0.40,0.2,0.02}
\definecolor{MyPurple}{RGB}{111,0,255}
\definecolor{MyRed}{rgb}{1.0,0.0,0.0}
\definecolor{MyGold}{rgb}{0.75,0.6,0.12}
\definecolor{MyDarkgray}{rgb}{0.66, 0.66, 0.66}

%% file: text/authors.tex
\author[a,1]{David Bau}
\author[a,b]{Jun-Yan Zhu} 
\author[c]{Hendrik Strobelt}
\author[d,e]{Agata Lapedriza}
\author[f]{Bolei Zhou}
\author[a]{Antonio Torralba}

\affil[a]{Computer Science and Artificial Intelligence Laboratory, Massachusetts Institute of Technology, Cambridge, MA 02139}
\affil[b]{Adobe Research, Adobe Inc., San Jose, CA 95110}
\affil[c]{Massachusetts Institute of Technology–International Business Machines (IBM) Watson Artificial Intelligence Laboratory, Cambridge, MA 02142}
\affil[d]{Media Lab, Massachusetts Institute of Technology, Cambridge, MA 02139}
\affil[e]{Estudis d’Informàtica, Multimèdia i Telecomunicació, Universitat Oberta de Catalunya, 08018 Barcelona, Spain}
\affil[f]{Department of Information Engineering, The Chinese University of Hong Kong, Shatin, Hong Kong SAR, China}

\leadauthor{Bau} 

\authorcontributions{}%
\authordeclaration{}%
\equalauthors{}%
\correspondingauthor{\textsuperscript{1}Corresponding author e-mail: davidbau@csail.mit.edu}%

%% file: text/abstract.tex
\begin{abstract}
Deep neural networks excel at finding hierarchical representations that solve complex tasks over large data sets. How can we humans understand these learned representations?
In this work, we present network dissection, an analytic framework to systematically identify the semantics of individual hidden units within image classification and image generation networks. First, we analyze a convolutional neural network (CNN) trained on scene classification and discover units that match a diverse set of object concepts. We find evidence that the network has learned many object classes that play crucial roles in classifying scene classes.  Second, we use a similar analytic method to analyze a generative adversarial network (GAN) model trained to generate scenes.  By analyzing changes made when small sets of units are activated or deactivated, we find that objects can be added and removed from the output scenes while adapting to the context.  
Finally, we apply our analytic framework to understanding adversarial attacks and to semantic image editing.
\end{abstract}

%% file: text/introduction.tex
\dropcap{C}an the individual hidden units of a deep network teach us how the network solves a complex task?  Intriguingly, within state-of-the-art deep networks, it has been observed that many single units match human-interpretable concepts that were not explicitly taught to the network: units have been found to detect objects, parts, textures, tense, gender, context, and sentiment~\cite{zhou2014object,zeiler2014visualizing,mahendran2015understanding,olah2018building,bau2018identifying,karpathy2016visualizing,radford2017learning}.
Finding such meaningful abstractions is one of the main goals of deep learning~\cite{bengio2013representation}, but the emergence and role of such concept-specific units is not well-understood.
Thus we ask: how can we quantify the emergence of concept units across the layers of a network?  What types of concepts are matched; and what function do they serve? When a network contains a unit that activates on trees, we wish to understand if it is a spurious correlation, or if the unit has a causal role that reveals how the network models its higher-level notions about trees.

To investigate these questions, we introduce \emph{network dissection}~\cite{zhou2018interpreting,bau2019gandissect}, our method for systematically mapping the semantic concepts found within a deep convolutional neural network.  The basic unit of computation within such a network is a learned convolutional filter; this architecture is the state-of-the-art for solving a wide variety of discriminative and generative tasks in computer vision~\cite{lecun1995convolutional,krizhevsky2012imagenet,simonyan2014very,goodfellow2014generative,vinyals2015show,he2016deep,isola2017image,zhu2017unpaired,karras2018progressive}. Network dissection identifies, visualizes, and quantifies the role of individual units in a network by comparing the activity of each unit to a range of human-interpretable pattern-matching tasks such as the detection of object classes.

Previous approaches for understanding a deep network include the use of salience maps~\cite{bach2015pixel,zhou2016learning,fong2017interpretable,lundberg2017unified,selvaraju2017grad,sundararajan2017axiomatic,smilkov2017smoothgrad,petsiuk2018rise}: those methods ask \emph{where} a network looks when it makes a decision.  The goal of our current inquiry is different: we ask \emph{what} a network is looking for, and why.  Another approach is to create simplified surrogate models to mimic and summarize a complex network's behavior~\cite{ribeiro2016should,kim2017tcav,koul2018learning}; and another technique is to train explanation networks that generate human-readable explanations of a network~\cite{hendricks2016generating}.  In contrast to those methods, network dissection aims to directly interpret the internal computation of the network itself, rather than training an auxiliary model.

We dissect the units of networks trained on two different types of tasks: image classification and image generation. In both settings, we find that a trained network contains units that correspond to high-level visual concepts that were not explicitly labeled in the training data. For example, when trained to classify or generate natural scene images, both types of networks learn individual units that match the visual concept of a `tree' even though we have  never taught the network the tree concept during training.

\input{figText/classifier-dissection.tex}
\input{figText/classifier-intervention.tex}
\input{figText/gan-dissection.tex}
\input{figText/gan-intervention.tex}

Focusing our analysis on the units of a network allows us to test the causal structure of network behavior by activating and deactivating the units during processing.  In a classifier, we use these interventions to ask whether the classification performance of a specific class can be explained by a small number of units that identify visual concepts in the scene class. For example, we ask how the ability of the network to classify an image as a ski resort is affected when removing a few units that detect snow, mountains, trees, and houses. Within a scene generation network, we ask how the rendering of objects in a scene is affected by object-specific units.  How does the removal of tree units affect the appearance of trees and other objects in the output image?

Finally, we demonstrate the usefulness of our approach with two applications.  We show how adversarial attacks on a classifier can be understood as attacks on the important units for a class.  And we apply unit intervention on a generator to enable a human user to modify semantic concepts such as trees and doors in an image by directly manipulating units.

%% file: figText/classifier-dissection.tex
\begin{figure*}[t]
\includegraphics[width=\textwidth]{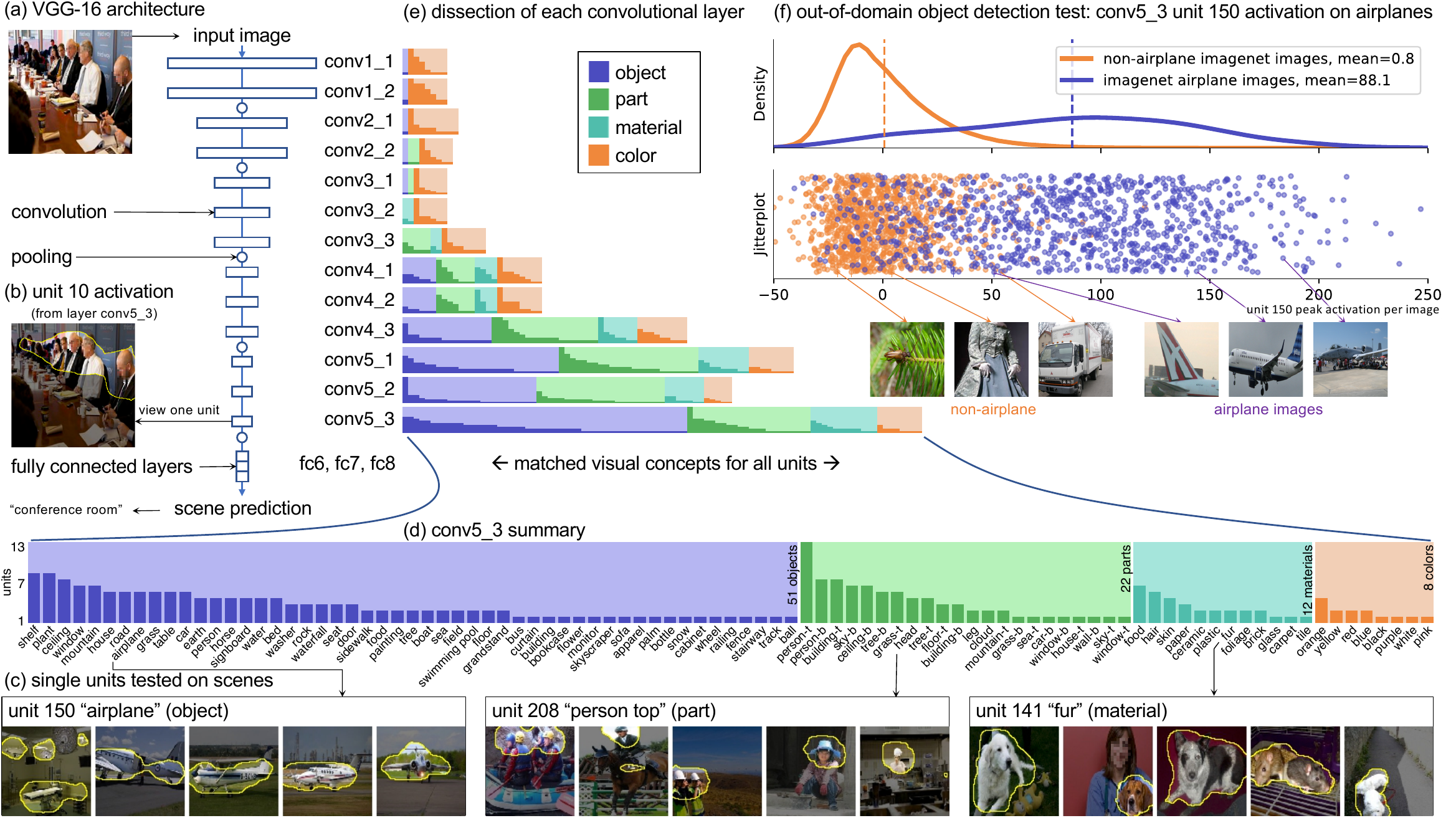}
\caption{The emergence of single-unit object detectors within a VGG-16 scene classifier.  (a) VGG-16 consists of 13 convolutional layers, \texttt{conv1\_1} through \texttt{conv5\_3}, followed by three fully connected layers, \texttt{fc6,7,8}.  (b) The activation of a single filter on an input image can be visualized as the region where the filter activates beyond its top 1\% quantile level.  (c) Single units are scored by matching high-activating regions against a set of human-interpretable visual concepts; each unit is labeled with its best-matching concept and visualized with maximally-activating images.  (d) Concepts that match units in the final convolutional layer are summarized, showing a broad diversity of detectors for objects, object parts, materials, and colors.  Many concepts are associated with multiple units. (e) Comparing all the layers of the network reveals that most object detectors emerge at the last convolutional layers.  (f) Although the training set contains no object labels, unit 150 emerges as an `airplane' object detector that activates much more strongly on airplane objects than non-airplane objects, as tested against a dataset of labeled object images not previously seen by the network.  The jitter plot shows peak activations for the unit on randomly sampled 1{,}000 airplane and 1{,}000 non-airplane Imagenet images, and the curves show the kernel density estimates of these activations.}
\lblfig{classifier-dissection}
\end{figure*}

%% file: figText/classifier-intervention.tex
\begin{SCfigure*}
\includegraphics[width=0.69\textwidth]{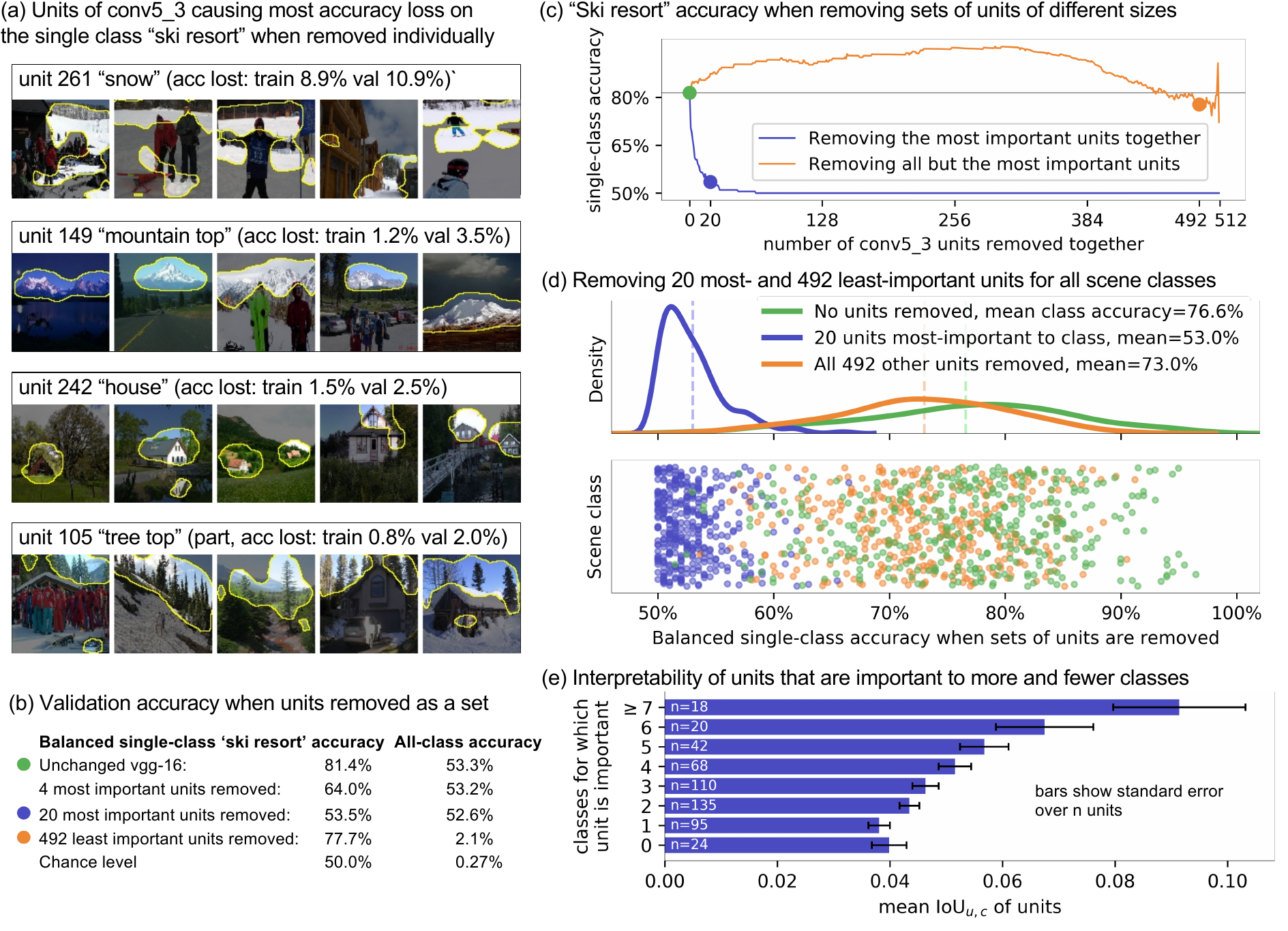}\vspace{-3pt}%
\caption{%
 A few units play important roles in classification performance.
(a) The four \texttt{conv5\_3} units cause the most damage to balanced classification accuracy for ``ski resort'' when each unit is individually removed from the network; dissection reveals that these most-important units detect visual concepts that are salient to ski resorts. (b) When the most-important units to the class are removed all together, balanced single-class accuracy drops to near chance levels. When the 492 least-important units in \texttt{conv5\_3} are removed all together (leaving only the 20 most-important units) accuracy remains high. (c) The effect on "ski resort" prediction accuracy when removing sets of units of successively larger sizes. These units are sorted in ascending and descending order of individual unit's impact on accuracy. (d) Repeating the experiment for each of 365 scene classes. Each point plots single-class classification accuracy in one of three settings: the original network; the network after removing the 20 units most-important to the class; and with all \texttt{conv5\_3} units removed except the 20 most-important ones. On the $y$ axis, classes are ordered alphabetically.  (e) The relationship between unit importance and interpretability.  Units that are among the top-4 important units for more classes are also closer matches for semantic concepts with as measured by $\IoU_{u,c}$.}
\lblfig{classifier-intervention}
\end{SCfigure*}

%% file: figText/gan-dissection.tex
\begin{figure*}[t]
\includegraphics[width=\textwidth]{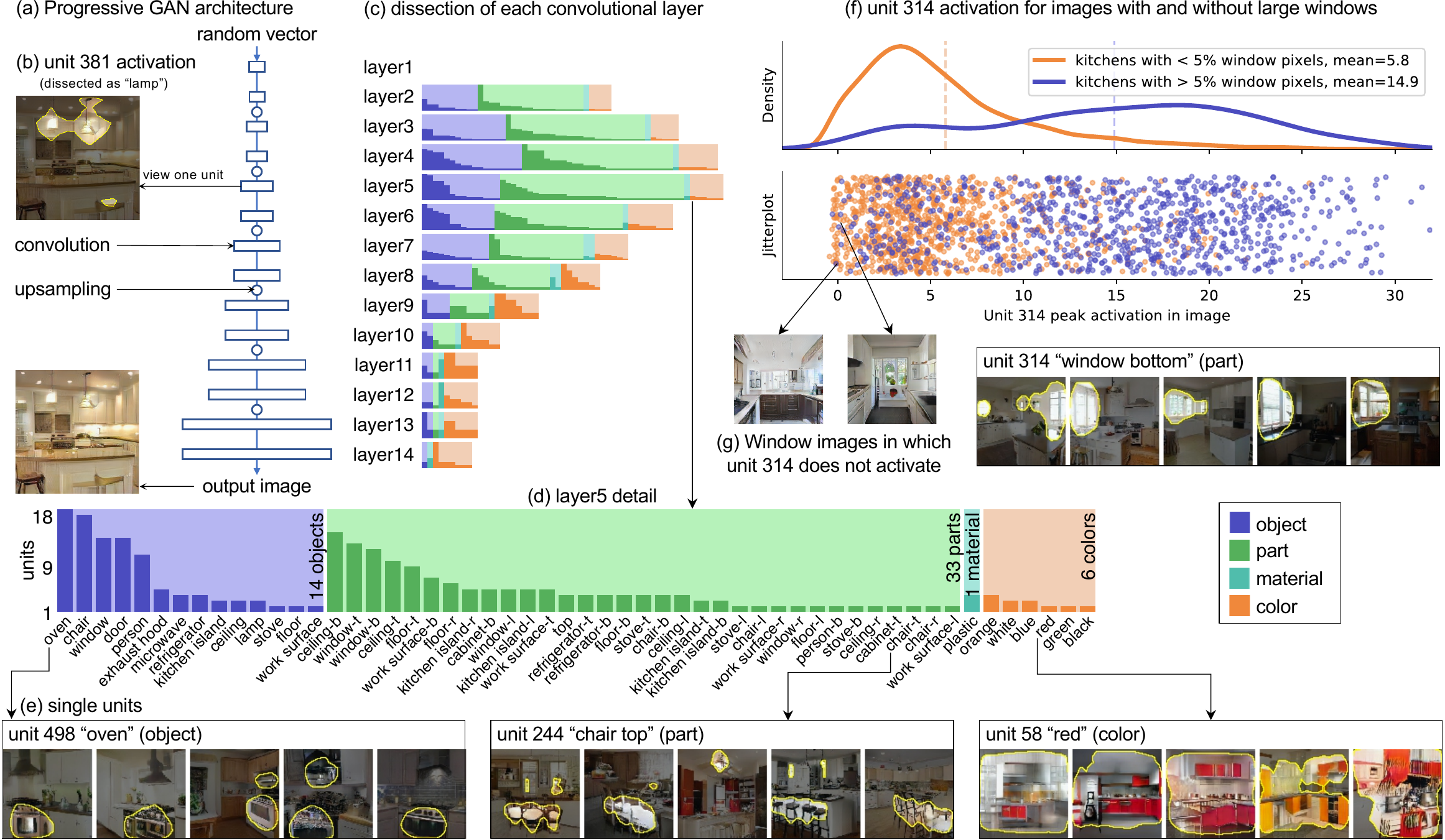}
\caption{The emergence of object- and part-specific units within a Progressive GAN generator`\cite{karras2018progressive}. (a) The analyzed Progressive GAN consists of 15 convolutional layers that transform a random input vector into a synthesized image of a kitchen.  (b) A single filter is visualized as the region of the output image where the filter activates beyond its top 1\% quantile level; note that the filters are all precursors to the output.  (c) Dissecting all the layers of the network shows a peak in object-specific units at \texttt{layer5} of the network. (d) A detailed examination of \texttt{layer5} shows more part-specific units than objects, and many visual concepts corresponding to multiple units.  (e) Units do not correspond to exact pixel patterns: a wide range of visual appearances for ovens and chairs are generated when an oven or chair part unit are activated.  (f) When a unit specific to window parts is tested as a classifier, on average the unit activates more strongly on generated images that contain large windows than images that do not.  The jitter plot shows the peak activation of unit 314 on 800 generated images that have windows larger than 5\% of the image area as estimated by a segmentation algorithm, and 800 generated images that do not.  (g) Some counterexamples: images for which unit 314 does not activate but where windows are synthesized nevertheless.}
\lblfig{gan-dissection}
\end{figure*}

%% file: figText/gan-intervention.tex
\begin{SCfigure*}
\includegraphics[width=0.72\textwidth]{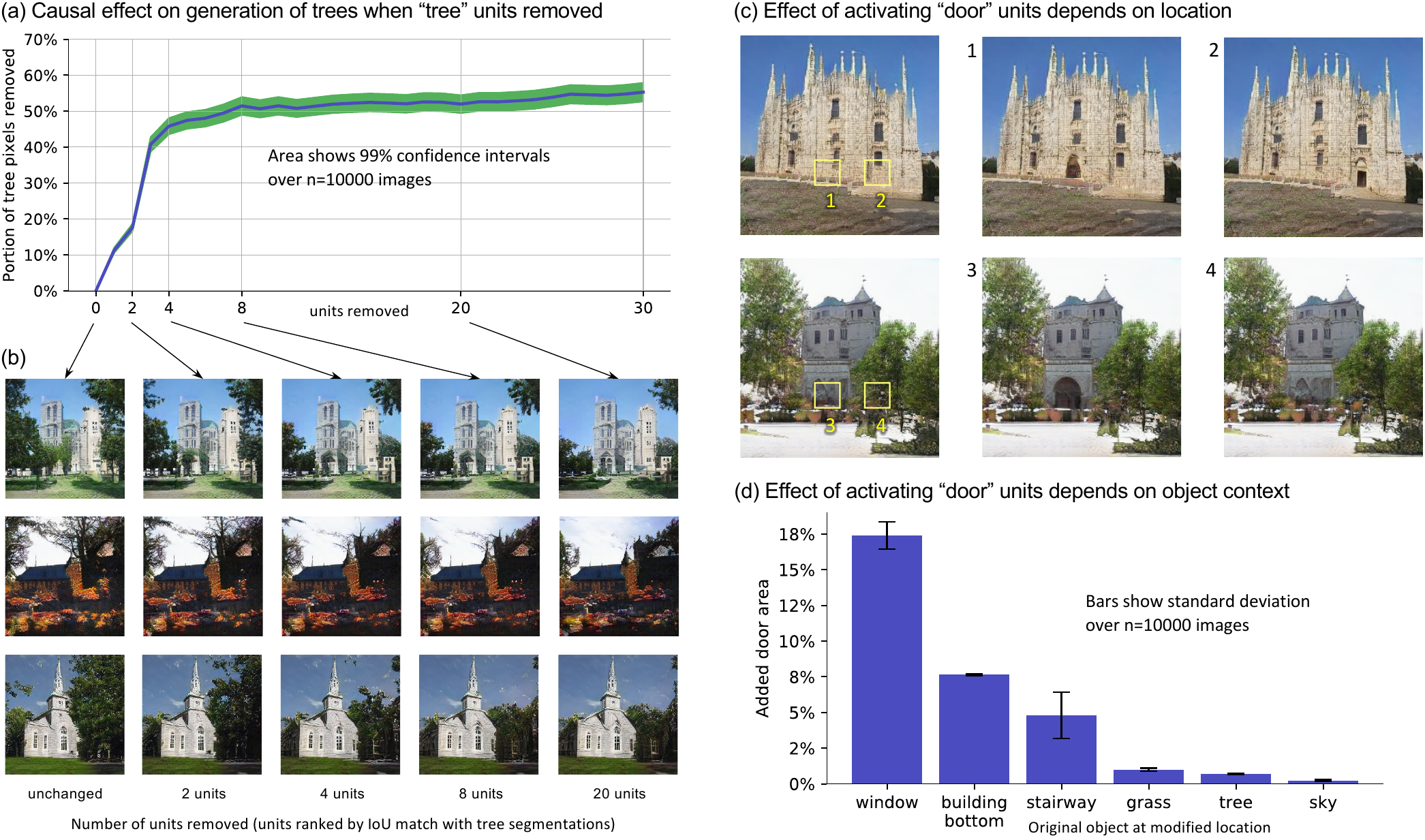}
\caption{The causal effect of altering units within a GAN generator. (a) When successively larger sets of units are removed from a GAN trained to generate outdoor church scenes, the tree area of the generated images is reduced. Removing 20 tree units removes more than half the generated tree pixels from the output.  (b) Qualitative results: removing tree units affects trees while leaving other objects intact.  Building parts that were previously occluded by trees are rendered as if revealing the objects that were behind the trees.  (c) Doors can be added to buildings by activating 20 door units.  The location, shape, size, and style of the rendered door depends on the location of the activated units.  The same activation levels produce different doors, or no door at all (case 4) depending on locations.  (d) Similar context dependence can be seen quantitatively: doors can be added in reasonable locations such as at the location of a window, but not in abnormal locations such as on a tree or in the sky.}

\lblfig{gan-intervention}
\end{SCfigure*}

%% file: text/results.tex
\section*{Results}

\subsection*{Emergence of Object Detectors in a Scene Classifier}

We first identify individual units that emerge as object detectors when training a network on a scene classification task. The network we analyze is a VGG-16 CNN~\cite{simonyan2014very} trained to classify images into 365 scene categories using the places365 data set~\cite{zhou2014learning}.  We analyze all units within the 13 convolutional layers of the network  (\reffig{classifier-dissection}a).  Please refer to materials and methods for further details on networks and datasets.

Each unit $u$ computes an activation function $a_u(x, p)$ that outputs a signal at every image position $p$ given a test image $x$. Filters with low-resolution outputs are visualized and analyzed at high-resolution positions $p$ using bilinear upsampling. Denote by $t_u$ the top 1\% quantile level for $a_u$, that is, writing $\Prob_{x, p}[\cdot]$ to indicate the probability that an event is true when sampled over all positions and images, we define the threshold $t_u \equiv \max_t \Prob_{x,p}[a_u(x, p) > t] \geq 0.01$.  In visualizations we highlight the activation region $\{p \;|\; a_u(x,p) > t_u\}$ above the threshold.  As seen in \reffig{classifier-dissection}b, this region can correspond to semantics such as the heads of all the people in the image. To identify filters that match semantic concepts, we measure the agreement between each filter and a visual concept $c$ using a computer vision segmentation model~\cite{xiao2018unified} $s_c: (x, p) \rightarrow \{0, 1\}$ that is trained to predict the presence of the visual concept $c$ within image $x$ at position $p$.  We quantify the agreement between concept $c$ and unit $u$ using the intersection over union (IoU) ratio:
\begin{align}
\IoU_{u,c} = \frac{\Prob_{x,p}[s_c(x, p) \land (a_u(x,p) > t_u)]}{\Prob_{x,p}[s_c(x, p) \lor (a_u(x,p) > t_u)]}
\end{align}
This IoU ratio is computed on the set of held-out validation set images. Within this validation set, each unit is scored against 1{,}825 segmented concepts $c$, including object classes, parts of objects, materials, and colors.  Then each unit is labeled with the highest-scoring matching concept.  \reffig{classifier-dissection}c shows several labeled concept detector units along with the five images with the highest unit activations.

When examining all 512 units in the last convolutional layer, we find many detected object classes and relatively fewer detected object parts and materials: \texttt{conv5\_3} units match 51 object classes, 22 parts, 12 materials, and 8 colors. Several visual concepts such as `airplane' and `head` are matched by more than one unit.  \reffig{classifier-dissection}d lists every segmented concept matching units in layer \texttt{conv5\_3} excluding any units with IoU ratio $< 4\%$, showing the frequency of units  matching each concept.  Across different layers, the last convolutional layer has the largest number of object classes detected by units, while the number of object parts peaks two layers earlier, at \texttt{conv5\_1}, which has units matching 28 object classes, 25 parts, 9 materials, and 8 colors (\reffig{classifier-dissection}e). A complete visualization of all the units of \texttt{conv5\_3} is provided in SI, as well as more detailed comparisons between layers of VGG-16, comparisons to layers of AlexNet~\cite{krizhevsky2012imagenet} and ResNet~\cite{he2016deep}, and an analysis of the texture versus shape sensitivity of units using a stylization method based on \cite{geirhos2018imagenet}.

Interestingly, object detectors emerge despite the absence of object labels in the training task.  For example, the aviation-related scene classes in the training set are `airfield', `airport terminal', `hangar', `landing deck', and `runway.' Scenes in these classes do not always contain airplanes, and there is no explicit `airplane' object label in the training set.  Yet unit 150 emerges as a detector that locates airplanes, scoring $\IoU=9.0\%$ agreement with our reference airplane segmentations in scene images.  The accuracy of the unit as an airplane classifier can be further verified on Imagenet~\cite{deng2009imagenet}, a dataset that contains 1{,}000 object classes; its images and classes are disjoint from the Places365 training set. Imagenet contains two airplane class labels: `airliner' and `warplane', and a simple threshold on unit 150 (peak activation $>23.4$) achieves 85.6\% balanced classification accuracy on the task of distinguishing these airplane classes from the other object classes.  \reffig{classifier-dissection}f shows the distribution of activations of this unit on a sample of airplane and non-airplane Imagenet images.

\subsection*{Role of Units in a Scene Classifier}
\hspace{1pt} How does the network use the above object detector units?  Studies of network compression have shown that many units can be eliminated from a network while recovering overall classification accuracy by retraining~\cite{wen2016learning,li2017pruning}. One way to estimate the importance of an individual unit is to examine the impact of the removal of the unit on mean network accuracy~\cite{morcos2018importance,zhou2018revisiting}.

To obtain a more fine-grained understanding of the causal role of each unit within a network, we measure the impact of removing each unit on the network's ability of classifying each individual scene class. Units are removed by forcing the specified unit to output zero and leaving the rest of the network intact. No retraining is done. 
Single-class accuracy is tested on the balanced two-way classification problem of discriminating the specified class from all the other classes.

The relationships between objects and scenes learned by the network can be revealed by identifying the most important units for each class.  For example, the four most important \texttt{conv5\_3} units for the class `ski resort' are shown in \reffig{classifier-intervention}a: these units damage ski resort accuracy most when removed. The units detect snow, mountains, houses, and trees, all of which seem salient to ski resort scenes.

To test whether the ability of the network to classify ski resorts can be attributed to just the most important units, we remove selected sets of units.  \reffig{classifier-intervention}b shows that removing just these 4 (out of 512) units reduces the network's accuracy at discriminating `ski resort' scenes from 81.4\% to 64.0\%, and removing the 20 most important units in \texttt{conv5\_3} reduces class accuracy further to 53.5\%, near chance levels (chance is 50\%), even though classification accuracy over all scene classes is hardly affected (changing from 53.3\% to 52.6\%, where chance is 0.27\%).  In contrast, removing the 492 least-important units, leaving only the 20 most important units in \texttt{conv5\_3}, has only a small impact on accuracy for the specific class, reducing ski resort accuracy by only 3.7\%, to 77.7\%.   Of course, removing so many units damages the ability of the network to classify other scene classes: removing the 492 least-important units reduces all-class accuracy to 2.1\% (chance is 0.27\%).

The effect of removing varying numbers of most-important and least-important units upon `ski resort' accuracy is shown in \reffig{classifier-intervention}c.  To avoid overfitting to the evaluation data, we rank the importance of units according to their individual impact on single-class ski resort accuracy on the training set, and the plotted impact of removing sets of units is evaluated on the held-out validation set.  The network can be seen to derive most of its performance for ski resort classification from just the most important units.  Single-class accuracy can even be improved by removing the least important units; this effect is further explored in SI.

This internal organization, in which the network relies on a small number of important units for most of its accuracy with respect to a single output class, is seen across all classes. \reffig{classifier-intervention}d repeats the same experiment for each of the 365 scene classes.  Removing the 20 most important \texttt{conv5\_3} units for each class reduces single-class accuracy to 53.0\% on average, near chance levels.  In contrast, removing the 492 least important units only reduces single-class accuracy by an average of 3.6\%, just a slight reduction.  We conclude that the emergent object detection done by units of \texttt{conv5\_3} is not spurious: each unit is important to a specific set of classes, and the object detectors can be interpreted as decomposing the network's classification of individual scene classes into simpler sub-problems.

Why do some units match interpretable concepts so well while other units do not?  The data in \reffig{classifier-intervention}e show that the most interpretable units are those that are important to many different output classes.  Units that are important to only one class (or none) are less interpretable, measured by \IoU.  We further find that important units are predominantly positively correlated with their associated classes, and different combinations of units provide support for each class.  Measurements of unit-class correlations and examples of overlapping combinations of important units are detailed in SI.

Does the emergence of interpretable units such as airplane, snow, and tree detectors depend on having training set labels that divide the visual world into hundreds of scene classes?  Perhaps the taxonomy of scenes encodes distinctions that are necessary to learn about objects. Or is it possible for a network to infer such concepts from the visual data itself?  To investigate this question, we next conduct a similar set of experiments on networks trained to solve unsupervised tasks.

\subsection*{Emergence of Object Detectors in a GAN}
\hspace{1pt} A generative adversarial network (GAN) learns to synthesize random realistic images that mimic the distribution of real images in a training set~\cite{goodfellow2014generative}. %
Architecturally, a trained GAN generator is the reverse of a classifier, producing a realistic image from a random input latent vector.  Unlike classification, it is an unsupervised setting: no human annotations are provided to a GAN, so the network must learn the structure of the images by itself.

Remarkably, GANs have been observed to learn global semantics of an image: for example, interpolating between latent vectors can smoothly transform the layout of a room~\cite{radford2015unsupervised} or change the texture of an object~\cite{zhu2016generative}. We wish to understand whether the GAN also learns hierarchical structure, for example, if it learns to decompose the generation of a scene into meaningful parts.

We test a Progressive GAN architecture~\cite{karras2018progressive} trained to imitate LSUN kitchen images~\cite{yu2015lsun}.  This network architecture consists of 15 convolutional layers, as shown in \reffig{gan-dissection}a.  Given a 512-dimensional vector sampled from a multivariate Gaussian distribution, the network produces a  256$\times$256 realistic image after processing the data through the 15 layers. As with a classifier network, each unit is visualized by showing the regions where the filter activates above its top 1\% quantile level, as shown in \reffig{gan-dissection}b.  Importantly, causality in a generator flows in the opposite direction as a classifier: when unit 381 activates on lamp shades in an image, it is not detecting objects in the image, because the filter activation occurs \emph{before} the image is generated.  Instead, the unit is part of the computation that ultimately renders the objects.

To identify the location of units in the network that are associated with object classes, we apply network dissection to the units of every layer of the network. In this experiment, the reference segmentation models and thresholds used are the same as those used to analyze the VGG-16 classifier. However, instead of analyzing agreement with objects that appear in the input data, we analyze agreement with segmented objects found in the generated output images.  As shown in \reffig{gan-dissection}c, the largest number of emergent concept units do not appear at the edge of the network as we saw in the classifier, but in the middle: the \texttt{layer5} has units that match the largest number of distinct object and part classes.

\reffig{gan-dissection}d shows each object, part, material, and color that matches a unit in \texttt{layer5} with $\IoU > 4\%$.  This layer contains 19 object-specific units, 41 units that match object parts, one material, and six color units.  As seen in the classification network, visual concepts such as `oven' and `chair' match many units. Different from the classifier, more object parts are matched than whole objects.

In \reffig{gan-dissection}d, individual units show a wide range of visual diversity:
the units do not appear to rigidly match a specific pixel pattern, but rather different appearances for a particular class, for example, various styles of ovens, or different colors and shapes of kitchen stools.

In \reffig{gan-dissection}f, we apply the window-specific unit 314 as an image classifier. We find a strong gap between the activation of the unit when a large window is generated and when no large window is generated.  Furthermore, a simple threshold (peak activation $>8.03$) can achieve a 78.2\% accuracy in predicting whether the generated image will have a large window or not.  Nevertheless, the distribution density curve reveals that images that contain large windows can be often generated without activating unit 314. Two such samples are shown in \reffig{gan-dissection}g.  These examples suggest that other units could potentially synthesize windows.

\subsection*{Role of Units in a GAN}
\hspace{1pt} The correlations between units and generated object classes are suggestive, but they do not prove that the units that correlate with an object class actually cause the generator to render instances of the object class.  To understand the causal role of a unit in a GAN generator, we test the output of the generator when sets of units are directly removed or activated.

We first remove successively larger sets of tree units from a Progressive GAN~\cite{karras2018progressive} trained on LSUN church scenes~\cite{yu2015lsun}.  We rank units in \texttt{layer4} according to $\IoU_{u,\text{tree}}$ to identify the most tree-specific units.  When successively larger sets of these tree units are removed from the network, the GAN generates images with fewer and smaller trees (\reffig{gan-intervention}a).  Removing the 20 most tree-specific units reduces the number of tree pixels in the generated output by 53.3\%, as measured over 10{,}000 randomly generated images.

When tree-specific units are removed, the generated images continue to look similarly realistic. Although fewer and smaller trees are generated, other objects such as buildings are unchanged.  Remarkably, parts of buildings that were occluded by trees are hallucinated, as if removing the trees reveals the walls and windows behind them (\reffig{gan-intervention}b).  The generator appears to have computed more details than are necessary to render the final output; the details of a building that are hidden behind a tree can only be revealed by suppressing the generation of the tree.  The appearance of such hidden details strongly suggests that the GAN is learning a structured statistical model of the scene that extends beyond %
a flat summarization of visible pixel patterns.

Units can also be forced on to insert new objects into a generated scene. 
We use $\IoU_{u,\text{door}}$ to find the 20 most door-specific units identified in \texttt{layer4} of the same outdoor church GAN.  
At tested locations, the activations for this set of 20 units are all forced to their high $t_u$ value. 
\reffig{gan-intervention}c shows the effect of applying this procedure to activate 20 door units at two different locations in two generated images. 
Although the same intervention is applied to all four cases, the doors obtained in each situation is different: In cases 1-3, the newly synthesized door has a size, style, and location that is appropriate to the scene context. In case 4, where door units are activated on a tree, no new door is added to the image.

\input{figText/adversarial.tex}

\reffig{gan-intervention}d quantifies the context-sensitivity of activating door units in different locations.  In 10{,}000 randomly generated images, the same 20-door-unit activation is tested at every featuremap location, and the number of newly synthesized door pixels is evaluated using a segmentation algorithm.  Doors can be easily added in some locations, such as in buildings and especially on top of an existing window, but it is nearly impossible to add a door into trees or in the sky.%

By learning to solve the unsupervised image generation problem, a GAN has learned units for emergent objects such as doors and trees. It has also learned a computational structure over those units that prevents it from rendering nonsensical output such as a door in the sky, or a door in a tree.

%% file: figText/adversarial.tex
\begin{figure*}[t]
\includegraphics[width=\textwidth,trim=0 8pt 0 0]{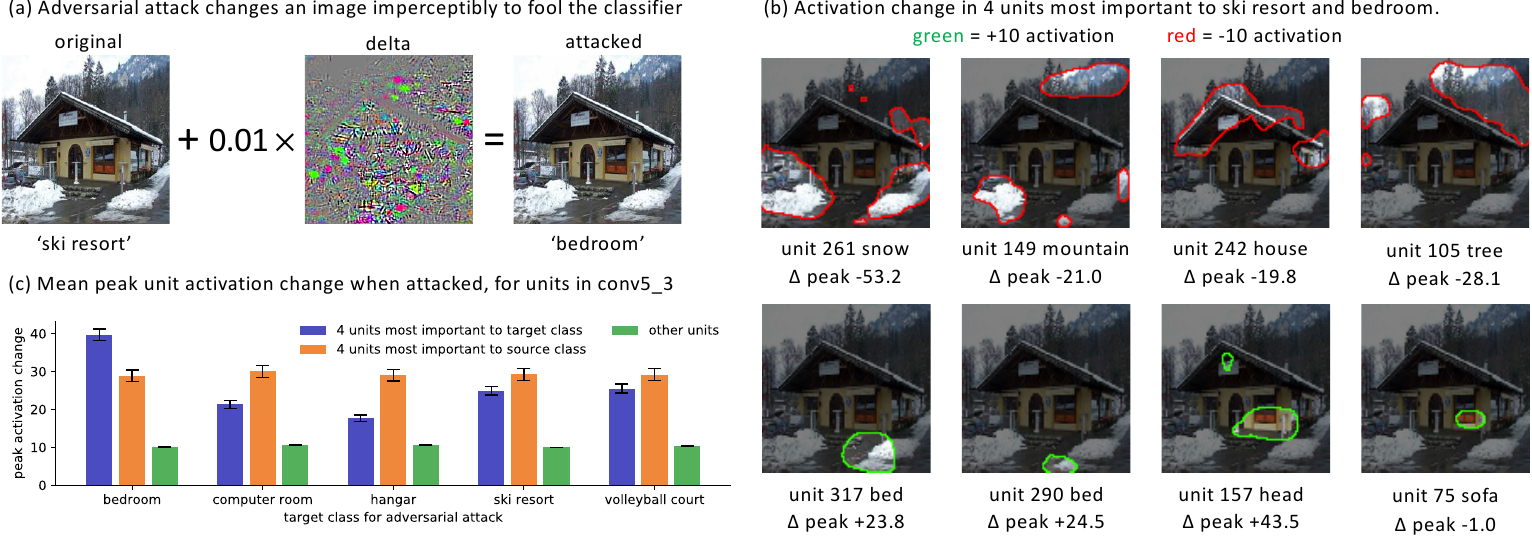}%
\caption{Application: visualizing an adversarial attack.  (a) the test image is correctly labeled as a ski resort, but when an adversarial perturbation is added, the visually indistinguishable result is classified as a bedroom.  (b) Visualization of the attack on the four most important units to the ski resort class and the four units most important to the bedroom class.  Areas of maximum increase and decrease are shown; $\Delta \text{peak}$ indicates the change in the peak activation level for the unit.  (c) Over 1000 images attacked to misclassify images to various incorrect target classes, the units that are changed most are those that dissection has identified as most important to the source and target classes.  Mean absolute value change in peak unit activation is graphed, with 99\% confidence intervals shown.}
\lblfig{adversarial}
\end{figure*}

%% file: text/applications.tex
\input{figText/ganpaint.tex}
\raggedbottom%

\section*{Applications}

We now turn to two applications enabled by our understanding of the role of units: understanding attacks on a classifier, and interactively editing a photo by activating units of a GAN.
\subsection*{Analyzing Adversarial Attack of a Classifier}
\hspace{1pt} The sensitivity of image classifiers to adversarial attacks is an active research area~\cite{szegedy2014intriguing,goodfellow2015explaining,carlini2017towards,madry2018towards}. To visualize and understand how an attack works, we can examine the effects on important object detector units.  In \reffig{adversarial}a, a correctly classified `ski resort' image is attacked to the target `bedroom' by the C \& W optimization method~\cite{carlini2017towards,rauber2017foolbox}.  The adversarial algorithm computes a small perturbation which, when added to the original, results in a misclassified image that is visually indistinguishable from the original image.  To understand how the attack works, we examine the four most important units to the ski resort class and the four most important units to the bedroom class.   \reffig{adversarial}b visualizes changes in the activations for these units between the original image and the adversarial image.  This reveals that the attack has fooled the network by reducing detection of snow, mountain, house, and tree objects, and by increasing activations of detectors for beds, person heads, and sofas in locations where those objects do not actually exist in the image.  \reffig{adversarial}c shows that, across many images and classes, the units that are most changed by an attack are the few units that are important to a class.

\subsection*{Semantic Paint using a GAN}
\hspace{1pt} Understanding the roles of units within a network allows us to create a human interface for controlling the network via direct manipulation of its units. We apply this method to a GAN to create an interactive painting application.  Instead of painting with a palette of colors, the application allows painting with a palette of high-level object concepts.  Each concept is associated with $20$ units that maximize $\IoU_{u, c}$ for the concept $u$.  \reffig{ganpaint}a shows our interactive interface.  When a user adds brush strokes with a concept, the units for the concept are activated (if the user is drawing) or zeroed (if the user is erasing).  \reffig{ganpaint}b shows typical results after the user adds an object to the image. The GAN deals with the pixel-level details of how to add objects while keeping the scene reasonable and realistic. Multiple changes in a scene can be composed for creative effects; movies of image editing demos are included in SI; online demos are also available at the website \url{http://gandissect.csail.mit.edu}.

%% file: figText/ganpaint.tex
\begin{figure}[t]
\includegraphics[width=\columnwidth]{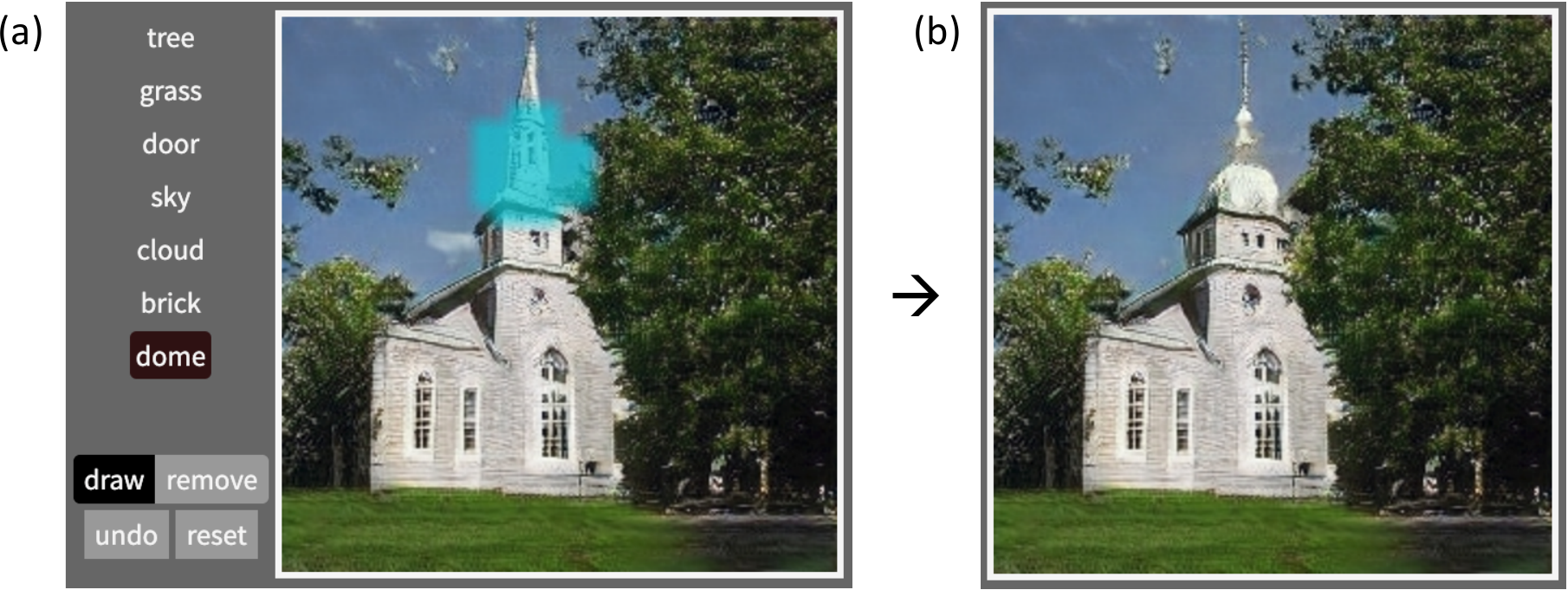}
\caption{Application: painting by manipulating GAN neurons.  (a) An interactive interface allows a user to choose several high-level semantic visual concepts and paint them on to an image.  Each concept corresponds to 20 units in the GAN. (b) After the user adds a dome in the specified location, the result is a modified image in which a dome has been added in place of the original steeple.  Once the user's high-level intent has been expressed by changing 20 dome units, the generator automatically handles the pixel-level details of how to fit together objects to keep the output scene realistic.}
\lblfig{ganpaint}
\end{figure}

%% file: text/discussion.tex
\section*{Discussion}
Simple measures of performance, such as classification accuracy, do not reveal \emph{how} a network solves its task: good performance can be achieved by networks that have differing sensitivities to shapes, textures, or perturbations~\cite{geirhos2018imagenet,ilyas2019adversarial}. 

To develop an improved understanding of how a network works, we have presented a way to analyze the roles of individual network units.  In a classifier, the units reveal how the network decomposes the recognition of specific scene classes into particular visual concepts that are important to each scene class.  And within a generator, the behavior of the units reveals contextual relationships that the model enforces between classes of objects in a scene.

Network dissection relies on the emergence of disentangled, human-interpretable units during training.  We have seen that many such interpretable units appear in state-of-the-art models, both supervised and unsupervised. How to train better disentangled models is an open problem that is the subject of ongoing efforts~\cite{chen2016infogan,higgins2017beta,zhang2018interpretable,achille2018emergence}.

We conclude that a systematic analysis of individual units can yield insights about the black box internals of deep networks.  By observing and manipulating units of a deep network, it is possible to understand the structure of the knowledge that the network has learned,
and to build systems that help humans interact with these powerful models.

%% file: text/methods.tex
\matmethods{\vspace{-18pt}%

\subsection*{Data sets}
Places365~\cite{zhou2017scene} consists of 1.80 million photographic images, each labeled with one of 365 scene classes.  The dataset also includes 36{,}500 labeled validation images (100 per class) that are not used for training.  Imagenet~\cite{deng2009imagenet} consists of 1.28 million photographic images, each focused on a single main object and labeled with one of 1{,}000 object classes. LSUN is a dataset with a large number of 256$\times$256 images in a few classes~\cite{yu2015lsun}.  LSUN kitchens consists of 2.21 million indoor kitchen photographs, and LSUN outdoor churches consists of 1.26 million photographs of church building exteriors.  Recognizable people in dataset images have been anonymized by pixelating faces in visualizations.

\subsection*{Tested Networks}
We analyze the VGG-16 classifier~\cite{simonyan2014very} trained by the Places365 authors~\citep{zhou2014learning} to classify Places365 images.  The network achieves classification accuracy of 53.3\% on the held-out validation set (chance is 0.27\%).  The 13 convolutional layers of VGG-16 are divided into 5 groups.  The layers in the first group contain 32 units that process image data at the full 224$\times$224 resolution; at each successive group the feature depth is doubled and the feature maps are pooled to halve the resolution, so that at the final stage that includes \texttt{conv5\_1} and \texttt{conv5\_3} the layers contain $512$ units at 14$\times$14 resolution.
The GAN models we analyze are trained by the Progressive GAN authors~\cite{karras2018progressive}.  The models are configured to generate 256$\times$256 output images using 15 convolutional layers divided into 8 groups, starting with 512 units in each layer at 4$\times$4 resolution, and doubling resolution at each successive group, so that \texttt{layer4} has 8$\times$8 resolution and 512 units and \texttt{layer5} has 16$\times$16 resolution and 512 units.  Unit depth is halved in each group after \texttt{layer6}, so that the 14th layer has 32 units and 256$\times$256 resolution.  The 15th layer (not pictured in \reffig{gan-dissection}a) produces a 3-channel RGB image.

\subsection*{Reference Segmentation}
To locate human-interpretable visual concepts within large-scale data sets of images, we use the Unified Perceptual Parsing image segmentation network~\citep{xiao2018unified} trained on the ADE20K scene dataset~\citep{zhou2017scene} and an assignment of rgb values to color names~\citep{van2009learning}.  The segmentation algorithm achieves mean IoU of 23.4\% on objects, 28.8\% on parts, and 54.2\% on materials.  To further identify units that specialize in object parts, we expand each object class into four additional object part classes which denote the top, bottom, left, or right half of the bounding box of a connected component.  Our reference segmentation algorithm can detect 335 object classes, 1452 object parts, 25 materials, and 11 colors.

\subsection*{Data Availability}
The code, trained model weights and data sets needed to reproduce the results in this paper are public and available for download at GitHub at \url{https://github.com/davidbau/dissect} and at the project
website \url{https://dissect.csail.mit.edu/}.
}

\showmatmethods{} %

%% file: text/acks.tex
\acknow{We are indebted to Aditya Khosla, Aude Oliva, William Peebles, Jonas Wulff, Joshua B. Tenenbaum, and William T. Freeman for their advice and collaboration.  And we are grateful for the support of the MIT-IBM Watson AI Lab, the DARPA XAI program FA8750-18-C0004, NSF 1524817 on Advancing Visual Recognition with Feature Visualizations, NSF BIGDATA 1447476, Grant RTI2018-095232-B-C22 from the Spanish Ministry of Science, Innovation and Universities to AL, Early Career Scheme (ECS) of Hong Kong (No.24206219) to BZ, and a hardware donation from NVIDIA.}

\showacknow{} %